\pdfoutput=1

\documentclass[11pt, dvipsnames]{article}

\usepackage[]{EMNLP2022}

\usepackage{times}
\usepackage{latexsym}
\usepackage{amsmath, amssymb, amsthm}
\usepackage[T1]{fontenc}
\usepackage{float}
\usepackage{colortbl}
\usepackage[utf8]{inputenc}

\usepackage{microtype}
\usepackage{bm}
\usepackage{soul, color}
\usepackage{graphicx}
\usepackage{caption}
\usepackage{subcaption}
\usepackage{array}
\usepackage{hhline}
\usepackage{amsfonts}
\usepackage{multirow}
\usepackage{booktabs}
\usepackage{adjustbox}
\usepackage{xcolor}
\usepackage{makecell}
\usepackage[most]{tcolorbox}
\usepackage{url}
\usepackage{tabularx}

\colorlet{LightLavender}{Lavender!30!}
\colorlet{LightRoyalBlue}{RoyalBlue!20!}
\colorlet{LightGray}{Gray!40!}
\colorlet{LightOrange}{YellowOrange!20!}
\tcbset{on line, 
    boxsep=1.0pt, left=0pt, right=0pt,top=0pt,bottom=0pt,
    colframe=white, colback=LightRoyalBlue,  
    highlight math style={enhanced}
}

\DeclareMathOperator*{\argmax}{arg\,max}

\newcommand\Tstrut{\rule{0pt}{2.3ex}}         
\newcommand\Bstrut{\rule[-1.3ex]{0pt}{0pt}}   

\title{Learning with Rejection for Abstractive Text Summarization}

\author{Meng Cao \qquad Yue Dong \qquad Jingyi He \qquad Jackie Chi Kit Cheung \\ \\
    Mila / McGill University, Montr\'{e}al, QC, Canada \\ \\
    {\small \{\tt meng.cao@mail, yue.dong2@mail, jingyi.he@mail, jcheung@cs\}.mcgill.ca}}

\begin{document}
\maketitle
\begin{abstract}
State-of-the-art abstractive summarization systems frequently hallucinate content that is not supported by the source document, mainly due to noise in the training dataset.
Existing methods opt to drop the noisy samples or tokens from the training set entirely, reducing the effective training set size and creating an artificial propensity to copy words from the source. 
In this work, we propose a training objective for abstractive summarization based on \emph{rejection learning}, in which the model learns whether or not to reject potentially noisy tokens. We further propose a regularized decoding objective that penalizes non-factual candidate summaries during inference by using the rejection probability learned during training.
We show that our method considerably improves the factuality of generated summaries in automatic and human evaluations when compared to five baseline models, and that it does so while increasing the abstractiveness of the generated summaries. \footnote{\url{https://github.com/mcao516/rej-summ}}
\end{abstract}

\section{Introduction}
Abstractive summarization has progressed significantly in  recent years, as evidenced by automatic evaluation scores like ROUGE \cite{lin-2004-rouge}. However, existing abstractive summarization systems are prone to produce non-factual summaries, which contain information that is not supported by the source \cite{kryscinski-etal-2019-neural, maynez-etal-2020-faithfulness, durmus-etal-2020-feqa, zhao-etal-2020-reducing, kryscinski-etal-2020-evaluating, narayan-etal-2021-planning}. According to recent studies, one cause of this phenomenon is noise in the training set, such as unsupported facts in the references \cite{kang-hashimoto-2020-improved, goyal-durrett-2021-annotating, raunak-etal-2021-curious}. Specifically, the widely-used negative log likelihood objective aims to match the underlying data distribution. 
If summarization models are trained on noisy datasets containing erroneous references, they will tend to imitate and even amplify undesired generations \cite{kang-hashimoto-2020-improved}.

Different approaches have been proposed to reduce non-factual hallucinations. \citet{falke-etal-2019-ranking, zhao-etal-2020-reducing} employ re-ranking based approaches to up-rank candidate summaries based on their factuality scores from out-of-the-box factual consistency evaluation systems.
\citet{kang-hashimoto-2020-improved, goyal-durrett-2021-annotating, goyal-etal-2022-training} use loss truncation-based method that drops high loss sentences/tokens during training. \citet{cao-etal-2022-hallucinated} propose reinforcement learning (RL)-based methods that penalize hallucinated tokens with negative rewards.

One instance of the reference summary containing unsupported information is shown in  Table~\ref{tab:example}. We can see that the reference summary contains details on the precise amount of government funding (\pounds 12.8m), yet this amount is absent from the source document.
This kind of noise in the reference summary is highly deceptive and could cause the summarization model to hallucinate. Furthermore, the noise can ``break'' the beam search process since the phrases associated with the noise (e.g., \emph{has been awarded [...]}) are decoded with high probability during inference. Therefore, we think it is crucial for the summarization model to learn what information is unknown and reject contexts in which hallucination is likely to happen.

\begin{table*}[t]
\renewcommand{\arraystretch}{1.05}
\centering
\small
\begin{tabular}{|p{15.5cm}|}
  \hline
  \Tstrut{\bf Source}: \\
    (...) Chris Cox, the university's director of development, said the research centre would translate discoveries made in the laboratory into new treatments. He said it would house 150 additional researchers who will be developing new ideas and treatments. Research will focus on radiation therapy, lung cancer, women's cancers, melanoma and haematological oncology. The centre is the result of a partnership between The University of Manchester, The Christie NHS Foundation Trust and Cancer Research UK. The Christie's chief executive Caroline Shaw said the funding would ``help facilitate groundbreaking research right here in Manchester''. (...) \Bstrut \\
  \hline
  \Tstrut{\bf Reference}: \\
  Plans for a new cancer research centre in Manchester have received a \emph{\pounds 12.8m} funding boost from the government. \Bstrut \\
  \hline
  \Tstrut{\bf BART (beam size=6)}: \\
    The University of Manchester has been awarded \tcbox[colback=LightOrange]{\pounds 3.5m} to help fund a new cancer research centre. \\
    The University of Manchester has been awarded \tcbox[colback=LightOrange]{\pounds 1.3m} to help fund a new cancer research centre. \\
    The University of Manchester has been awarded \tcbox[colback=LightOrange]{\pounds 1.2m} to help fund a new cancer research centre. \\
    The University of Manchester has been awarded \tcbox[colback=LightOrange]{\pounds 1.5m} to help fund a new cancer research centre. \\
    The University of Manchester has been awarded \tcbox[colback=LightOrange]{\pounds 5m} to help fund a new cancer research centre. \\
    The University of Manchester has been awarded \tcbox[colback=LightOrange]{\pounds 4.5m} to help fund a new cancer research centre. \Bstrut \\
  \hhline{=}
  \Tstrut{\bf BART trained with rejection (greedy decoding)}: \\
  The University of Manchester has been awarded \tcbox[colback=LightOrange]{\pounds{\texttt{<REJ>}}m} to help fund its new cancer research centre. \Bstrut \\
  \hline
  \Tstrut{\bf BART trained with rejection (regularized decoding)}: \\
  Details of a new cancer research centre at the University of Manchester have been revealed. \Bstrut \\
  \hline
\end{tabular}
\caption{\label{tab:example} Example of a BBC article from \textsc{XSum} test set. In this example, the source document does not have information on the specific value of the government funding. However, BART model \cite{lewis-etal-2020-bart} trained using the MLE objective hallucinates different funding values in its generation (highlighted). In the bottom, we show that the BART model trained using the rejection objective (see Section \ref{sec:learning_with_rej}) work successfully reject the hallucinated token during inference. After applying the regularized decoding (Section \ref{sec:regularzied_decoding}), the model generates a factual summary without hallucinated information.
}
\end{table*}

In this work, we propose a training objective that enables  the summarization model to recognize and reject unsupported target text spans. Inspired by rejection learning for image classification 
\cite{NIPS2008_3df1d4b9, 9115b6c49f784e5191a2228055640de0, pmlr-v97-thulasidasan19a}, we introduce a new rejection class at training time that offers the model an option to reject confusing tokens rather than assigning high probability to all reference tokens. Besides, our method does not require labeling or data preprocessing.
The learning is based on the training dynamics of abstractive summarization systems where clean samples are learned quickly and noisy samples have relatively high loss \cite{goyal-etal-2022-training, kang-hashimoto-2020-improved}.
We demonstrate that the rejection learning objective allows the summarization model to learn features that are indicative of unsupported information.
As shown in the example in Table \ref{tab:example}, the model trained with the rejection objective correctly rejects after the phrase ``\emph{has been awarded}'' since such information is absent from the source. According to analysis on human-annotated factuality datasets, our model can accurately reject 77.6\% of non-factual hallucinations generated by state-of-the-art abstractive summarization model.

Furthermore, we also provide a regularized decoding objective for summarization models trained with the rejection training objective.
We add a regularization term to the original decoding objective to penalize candidates using the rejection probability learned during training.
We demonstrate that the regularized decoding objective, combined with rejection training, can vastly improve the factuality of summarization models using both automatic and human evaluation. Compared with the MLE baseline, our method improves the sentence factuality rate by 25.4\% ($34.6\%\rightarrow43.4\%$, measured by DAE \cite{goyal-durrett-2021-annotating}) and $41.7\%$ ($22.8\%\rightarrow32.3\%$, measured by \textsc{FactCC} \cite{kryscinski-etal-2020-evaluating}). A recent study has shown that part of the improvement of faithfulness brought by prior methods stems from an increased level of extractiveness of the model outputs \cite{ladhak-etal-2022-faithful}. We demonstrate that our method improves model's factuality while simultaneously increasing the abstractiveness of the generated summaries. 

Our main contributions in this work are:
(\textit{i}) We introduce a rejection learning objective for training abstractive summarization models. We demonstrate how the additional training objective enables the summarization model to precisely reject hallucinated tokens during inference (Section \ref{sec:rej_analysis}). (\textit{ii}) We propose a novel factuality-aware regularized decoding objective based on the rejection probability learned during training. Our method significantly improves the factuality of generated summaries in automatic and human evaluations when compared to five baseline models. (\textit{iii}) We demonstrate that, in comparison to the baselines, the summaries produced by our method also exhibit a greater level of abstractiveness with more novel $n$-grams and less copying. 





\section{Related Work}
With the advent of transformer-based summarization models \citep{lewis-etal-2020-bart,  zhang2020pegasus,qi-etal-2020-prophetnet,liu-liu-2021-simcls,liu-etal-2022-brio}, current system-generated summaries are often fluent and salient.  However, detailed analysis of these systems' outputs reveals that the majority of the generation contains hallucinations \citep{kryscinski-etal-2020-evaluating,pagnoni-etal-2021-understanding}. Moreover, annotations from \citet{maynez-etal-2020-faithfulness} suggest that the majority of these hallucinations are extrinsic; namely, they contain information that is not directly inferable from the source. 

Many prior studies consider extrinsic hallucinations generated by models as undesirable, and propose different methods to reduce these hallucinations \citep{falke-etal-2019-ranking,dong-etal-2020-multi, zhu-etal-2021-enhancing,cao-etal-2020-factual, pagnoni-etal-2021-understanding,nan-etal-2021-entity}. On the other hand, some prior work \citep{maynez-etal-2020-faithfulness, cao-etal-2022-hallucinated, dong2022faithful} suggest many of these hallucinations are actually factual and can be useful to provide additional background information. 

The most similar work to ours is the approach of using loss truncation to reduce noise in the training stage. \citet{kang-hashimoto-2020-improved} and \citet{goyal-etal-2022-training} propose sentence and token-level loss truncation for hallucination reduction. These strategies are based on the assumption that noisy examples are difficult to learn, resulting in high training loss. Simply removing these difficult-to-learn high log loss samples might reduce model hallucination. However, dropping too many samples could make the training process inefficient. It also makes the summarization model more extractive.
Dropping high loss token might not be effective either since the model does not learn the right token to generate during decoding.

We instead use the so-called \textit{rejection} (or abstention) approach to combat noise in training. The loss abstention approach was proposed in \citet{pmlr-v97-thulasidasan19a} for classification tasks, where a classifier is trained to abstain on mislabeled images during training. Inspired by this idea, we introduce a rejection loss for generation tasks (Section~\ref{sec:learning_with_rej}). The advantage of loss abstention over loss truncation is its ability to learn text features that are likely to lead to hallucination rather than simply ignore them. Such features can be used during inference to avoid hallucinations.

\section{Method}
\subsection{Learning with Rejection} \label{sec:learning_with_rej}
\paragraph{Problem  Formulation} We consider abstractive summarization as a conditional language generation task. Given a source document $\mathbf{x}=(x_1, ... , x_L)$ with $L$ tokens, the task is to learn a probabilistic model $\hat{p}_{\theta}(\mathbf{y}|\mathbf{x})$ that generates a summary $\mathbf{y}=(y_1, ..., y_{|\mathbf{y}|})$, where $y_i$ comes from a pre-defined vocabulary $\mathcal{V}$. $\theta$ denotes the parameters of the model. $\hat{p}_{\theta}$ is modeled by decomposing the probability of a sequence into conditional probabilities of each token given the previous context: $\hat{p}_{\theta}(\mathbf{y})=\prod_{t=1}^{|y|} \hat{p}_{\theta}(y_t \ | \ \mathbf{y}_{<t}, \mathbf{x})$. 
Usually, $\hat{p}_{\theta}$ is trained using the maximum likelihood estimation (MLE) objective, which aims to maximize the likelihood of the human-written reference summaries:
\begin{equation}
\theta^* = \argmax_{\theta} \sum_{\mathbf{x}, \mathbf{y}^* \sim \mathcal{D}} \log{\hat{p}_{\theta}({\mathbf{y}}^{*} \mid \mathbf{x}; \theta)},
\end{equation}
where $\mathcal{D}$ denotes the training set and $\mathbf{y}^*$ is the reference summary. 
For a given document $\mathbf{x}$ and reference summary $\mathbf{y}^*$,
the MLE objective is equivalent to minimizing the sum of negative log-likelihoods of the ground-truth tokens:
\begin{equation}
\begin{split}
    \mathcal{L}(\theta) &= - \log{\hat{p}_{\theta}({\mathbf{y}}^{*} \mid \mathbf{x}; \theta)} \\
     &= - \sum_{t=1}^{|\mathbf{y}|} \log{\hat{p}_{\theta}({y}^{*}_{t} \mid \mathbf{x},  \mathbf{y}^*_{<t}; \theta)}, \\
\end{split}\label{eq:nll}
\end{equation}
where $y_i^* \in \mathcal{V}$ is the ground-truth token from the reference summary. Equation \ref{eq:nll} is also equivalent to the cross-entropy loss objective where the true distribution is a one-hot distribution.

For the sake of notational brevity, we define $p_t^i = {\hat{p}_{\theta}(y_t^i \mid \mathbf{y}^*_{<t}, \mathbf{x})}$, ($0 \leq i \leq |\mathcal{V}|-1$) as the conditional probability of the $i$-th vocabulary token at the $t$-th time-step given the ground-truth prefix.

\paragraph{The Rejection Loss} The MLE objective seeks to maximize the likelihood of all training samples. Knowing that the underlying training data is noisy, we want the model to automatically recognize unsupported text spans in references and reject them during training \cite{pmlr-v97-thulasidasan19a}. To do this, we can add an additional rejection class $r$ out of the original $|\mathcal{V}|$ classes (i.e., tokens). 
Let $p^{r}_t=\hat{p}_{\theta}(r \mid \mathbf{x}, \mathbf{y}^*_{<t})$ denote the probability of rejection at the $t$-th step, the modified loss function is
\begin{equation}
\resizebox{\columnwidth}{!}{
$
\begin{split}
&    \mathcal{L}(\theta) = \\
& -\sum_{t=1}^{|\mathbf{y}|} \Big [ (1 - p^{r}_t) \log{\frac{\hat{p}_{\theta}(y_t^* \mid \mathbf{x}, \mathbf{y}^*_{<t}; \theta)}{1 - p^{r}_t}} + \alpha \cdot \mathcal{R}_T(p^{r}_t) \Big ] \\
\end{split}\label{eq:obj}
$
}
\end{equation}
where $\mathcal{R}_T(p) = \log{(1-p)}$ is the regularization function for rejection. In the first term, increasing either $\hat{p}_{\theta}(y^* \mid \mathbf{x}, \mathbf{y}^*_{<t}; \theta)$ or $p^{r}_t$ will minimize the loss function.
During training, when the ground-truth token contains misannotations or unsupported information, term $\log{\hat{p}_{\theta}(y_t^* \mid \mathbf{x}, \mathbf{y}^*_{<t}; \theta)}$ will be abnormally high \cite{kang-hashimoto-2020-improved}. In this situation, minimizing the log loss results in a suboptimal model that hallucinates. By introducing the rejection class, the model has the alternative of increasing the probability of rejection in order to minimize the total loss.
In an extreme case where $p^{r}_t=1$, the first loss term will be zero. To prevent the model from deteriorating into the extreme case, we add a second regularization term that penalizes large rejection probability.
During training, we set $p^{r}_t=0$ to fully recover the original cross-entropy loss (Equation~\ref{eq:nll}) for the first $m$ steps. This allows the model to quickly absorb easy patterns.
In our experiments, we applied the rejection loss only to entity tokens.
 
\subsection{Decoding with Rejection}  \label{sec:regularzied_decoding}
Let's consider how to do decoding for a model trained with the rejection loss objective above.
At inference time, the rejection probability reflects the degree of uncertainty the model has about its prediction.
When the rejection class is assigned a high probability at time step $t$, the generated content is likely to be hallucinated and non-factual at this step (see Section \ref{sec:rej_analysis}).
Therefore, we can make use of the rejection probability to penalize non-factual candidate summaries.
We consider the following regularized decoding objective:
\begin{equation}
    \mathbf{y}^* = \argmax_{\mathbf{y} \in \mathcal{Y}} \big ( \log{\hat{p}_{\theta}(\mathbf{y} \mid \mathbf{x}; \theta)} - \lambda \cdot \mathcal{R}_D(\mathbf{y}) \big )
\label{eq:decoding_obj}
\end{equation}
where 
\begin{equation}\label{eq:sum_reg}
    \mathcal{R}_{D}(\mathbf{y}) = \frac{1}{|y|} \sum_{t=1}^{|y|} \log^k{\frac{1}{1-p^r_t}}
\end{equation}
and $\mathcal{Y}$ is the hypothesis space. $k$ is the exponent for the log that can amplify the effect of the regularization term when $p^r_t$ is large. We set $k=1$ in the following experiments.
We also consider a max regularizer that finds the token with the highest rejection probability in the generated summary: 
\begin{equation}\label{eq:max_reg}
    \mathcal{R}_{D, \mathrm{max}}(\mathbf{y}) = \max_{t=1}^{|y|} \log^k{\frac{1}{1-p^r_t}}
\end{equation}
During inference, the rejection token is only used for regularization and will not be generated. At each decoding step $t$, we only select tokens from the original vocabulary $\mathcal{V}$. The probability of each token $p_t^i$ is divided by $1-p_t^r$ to make sure the model outputs a proper probability distribution over vocabulary words.

\section{Experiments}
\subsection{Experimental Settings}
\paragraph{Datasets}
We evaluated our method on the \textsc{XSum} \cite{xsum-emnlp} dataset. \textsc{XSum} contains $226{,}711$ online British Broadcasting Corporation (BBC) articles covering domains like news,
politics, sports, business and more.
Each article is paired with a single sentence summary written by the BBC journalists. The dataset is split into three subsets: training ($204{,}045$, 90\%), validation ($11{,}332$, 5\%), and test ($11{,}334$, 5\%) sets.
We choose \textsc{XSum} because it is more abstractive than other summarization datasets. It also relatively noisy due to the way it is created \cite{maynez-etal-2020-faithfulness}.

For fine-grained entity-level factuality analysis (see Section~\ref{sec:rej_analysis}),
we use the \textsc{XEnt} dataset created by \citet{cao-etal-2022-hallucinated} . \textsc{XEnt} contains 800 article and summary pairs where the articles are sampled from \textsc{XSum} test set and the summaries are generated using a fine-tuned BART model \cite{lewis-etal-2020-bart}. In \textsc{XEnt}, each entity in the summary is annotated with three labels: non-hallucinated, factual hallucination, and non-factual hallucination. Factual hallucinations refer to an entity that is verifiable by world knowledge but not directly inferable from the source text. 
We evaluate our model on the test set of \textsc{XEnt} which contains 240 summaries and 835 entities.

\paragraph{Baselines} We compare our method with five baselines:
\textbf{(1)} State-of-the-art abstractive summarization model BART \cite{lewis-etal-2020-bart}. \textbf{(2)} The loss truncation method proposed by \citet{kang-hashimoto-2020-improved}. \textbf{(3)} The token-level loss truncation method proposed by \citet{goyal-etal-2022-training}. \textbf{(4)} GOLD (generation by off-policy learning from demonstrations), an off-policy RL algorithm proposed by \citet{pang2021text}. \textbf{(5)} A factuality-aware off-policy RL algorithm for text summarization proposed by \cite{cao-etal-2022-hallucinated}.

BART is a frequently used transformer-based model for summarization. It employs the encoder-decoder architecture, which is essentially a denoising autoencoder that reconstructs the original text with injected noise, such as token deletion, replacement, masking and sentence permutation.  The loss truncation method proposed by \citet{kang-hashimoto-2020-improved} adaptively discards high log loss examples as a way to reduce hallucination in generation. Similarly, the token-level loss truncation in \citet{goyal-etal-2022-training}  removes high-loss and low-loss tokens that are challenging or too easy to learn. GOLD  \cite{pang2021text} is proposed to alleviate the discrepancies between training and evaluation objectives in MLE (e.g, over-generalization, exposure bias). \citet{pang2021text} shows that GOLD outperforms BART on text summarization in terms of ROUGE scores and human evaluation. \citet{cao-etal-2022-hallucinated} augmented GOLD with factuality rewards generated using their entity-level factuality classifier.

\paragraph{Implementation Details}
In the following experiments, we use pre-trained \textsc{Bart-large} as the backbone for our model. The size of hidden layers is set to 1024 and the max sequence length is set to 1024 for both the encoder and decoder modules. We fine-tuned our model on the \textsc{XSum} training set for 20,000 steps with a warm-up period of 500 steps.
We use the Adam optimizer \cite{kingma2014adam} with $\epsilon=\textrm{1e-8}$ and an initial learning rate 3e-5.
Both dropout and attention dropout are applied in each layer with a dropout rate set to 0.1. The $\alpha$ in Equation~\ref{eq:obj} is set to 1.0 unless otherwise specified. We use the sum regularizer defined in Equation \ref{eq:sum_reg} for our main results.
All experiments are conducted on 4 Tesla V100 GPUs with 32GB of memory.

\paragraph{Automated Evaluation Metrics}
For sentence-level faithfulness evaluation, we use \textsc{FactCC} \cite{kryscinski-etal-2020-evaluating} and DAE \cite{goyal-durrett-2021-annotating}. \textsc{FactCC} is a sentence-level faithfulness classifier trained on synthetic dataset that is created based on analysis of errors made by state-of-the-art summarization models. DAE is trained on fine-grained, human annotations of factual errors made by abstractive summarization models.
Given a document and summary pair $(\mathbf{x}, \mathbf{y})$, both methods predict a label $y \in \{0, 1\}$ where 0 indicts the summary being non-faithful and 1 being faithful. The percentage of sentences that are predicted as faithful by the classifier is used as the sentence factuality rate.

For entity-level factuality evaluation, we use \textsc{EntFA} proposed by \citet{cao-etal-2022-hallucinated}. \textsc{EntFA} is a $k$-NN model that uses prior and posterior probabilities as features to predict the factuality and hallucination status of the entities in a given summary. We use the percentage of entities in the output summaries that are predicted as \emph{non-factual hallucinations} as the entity hallucination rate in Table \ref{tab:auto_eval}.
For all three evaluation methods, we use the evaluation scripts and parameters provided in the papers.
To measure the impact of each method on summary abstractiveness, we calculate the percentage of novel $n$-grams and \emph{coverage}, the percentage of words in the summary that are part of an extractive fragment of the source document \cite{grusky-etal-2018-newsroom}

\subsection{Factuality Evaluation}
In this section, we conduct automatic and human evaluation on summary factuality.
\subsubsection{Automatic Evaluation Results}
\label{sec:automatic_eval}
\setlength\heavyrulewidth{0.25ex}
\begin{table*}[t!]
\renewcommand{\arraystretch}{0.95}
\centering
\begin{adjustbox}{width=\textwidth, center}
\begin{tabular}{l|rr|r|rr|r}
\toprule
\multirow{2}{*}{\bf Model} & \multicolumn{2}{c|}{\bf Sentence Factuality Rate} & {\bf Entity Hal. Rate} & \multicolumn{2}{c|}{\bf Novel $n$-gram} & \multirow{2}{*}{\bf ROUGE-1} \\ 
~ & {\small \textsc{FactCC} $\uparrow$} & {\small DAE $\uparrow$} & {\small \textsc{EntFA} $\downarrow$} & {\small Unigram $\uparrow$} & {\small Bigram $\uparrow$} & ~ \\ 
\midrule
\multicolumn{7}{c}{Sentence-level Loss Truncation} \\
\midrule
MLE$^\ddagger$ & 22.8\% & 34.6\% & 17.2\% & 27.9\% & 74.5\% & 45.1 \\
\ +SLT$^\dagger$ ($\textrm{c}=0.3$) & $\tcbhighmath{\uparrow 4.8\%}$ 23.9\% & $\tcbhighmath{\uparrow 5.2\%}$ 36.4\% & $\tcbhighmath{\downarrow 6.4\%}$ 16.1\% & $\tcbhighmath[colback=LightOrange]{\downarrow 3.9\%}$ 26.8\% & $\tcbhighmath[colback=LightOrange]{\downarrow 1.5\%}$ 73.4\% & $\tcbhighmath[colback=LightOrange]{\downarrow 2.2\%}$ 44.1 \\
\ +SLT$^\dagger$ ($\textrm{c}=0.5$) & $\tcbhighmath{\uparrow 7.5\%}$ 24.5\% & $\tcbhighmath{\uparrow 7.7\%}$ 37.5\% & $\tcbhighmath{\downarrow 4.7\%}$ 16.4\% & $\tcbhighmath[colback=LightOrange]{\downarrow 4.3\%}$ 26.7\% & $\tcbhighmath[colback=LightOrange]{\downarrow 1.9\%}$ 73.1\% & $\tcbhighmath[colback=LightOrange]{\downarrow 3.1\%}$ 43.7 \\
\ +SLT$^\dagger$ ($\textrm{c}=0.7$) & $\tcbhighmath{\uparrow 11.8\%}$ 25.5\% & $\tcbhighmath{\uparrow 12.1\%}$ 38.8\% & $\tcbhighmath{\downarrow 7.6\%}$ 15.9\% & $\tcbhighmath[colback=LightOrange]{\downarrow 4.7\%}$ 26.6\% & $\tcbhighmath[colback=LightOrange]{\downarrow 1.7\%}$ 73.2\% & $\tcbhighmath[colback=LightOrange]{\downarrow 5.3\%}$ 42.7 \\
\midrule
\multicolumn{7}{c}{Token-level Loss Truncation} \\
\midrule
\ +TLT$^\dagger$ ($\textrm{c}=0.3$) & $\tcbhighmath{\uparrow 11.4\%}$ 25.4\% & $\tcbhighmath{\uparrow 0.3\%}$ 34.7\% & $\tcbhighmath{\downarrow 3.5\%}$ 16.6\% & $\tcbhighmath{\uparrow 0.4\%}$ 28.0\% & $\tcbhighmath{\uparrow 1.6\%}$ 75.7\% & $\tcbhighmath[colback=LightOrange]{\downarrow 4.0\%}$ 43.3 \\
\ +TLT$^\ddagger$ ($\textrm{c}=0.5$) & $\tcbhighmath{\uparrow 3.0\%}$ 23.5\% & $\tcbhighmath{\uparrow 5.8\%}$ 36.6\% & $\tcbhighmath[colback=LightOrange]{\uparrow 3.5\%}$ 17.8\% & $\tcbhighmath[colback=LightOrange]{\downarrow 9.3\%}$ 25.3\% & $\tcbhighmath[colback=LightOrange]{\downarrow 1.6\%}$ 73.3\% & $\tcbhighmath[colback=LightOrange]{\downarrow 6.4\%}$ 42.2 \\
\ +TLT$^\dagger$ ($\textrm{c}=0.7$) & $\tcbhighmath{\uparrow 19.3\%}$ 27.2\% & $\tcbhighmath{\uparrow 20.5\%}$ 41.7\% & $\tcbhighmath[colback=LightOrange]{\uparrow 15.7\%}$ 19.9\% & $\tcbhighmath[colback=LightOrange]{\downarrow 14.0\%}$ 24.0\% & $\tcbhighmath[colback=LightOrange]{\downarrow 4.3\%}$ 71.3\% & $\tcbhighmath[colback=LightOrange]{\downarrow 21.3\%}$ 35.5 \\
\midrule 
\multicolumn{7}{c}{RL with Factuality Rewards} \\
\midrule
RL$^\ast$ & 22.1\% & 33.3\% & 17.8\% & 28.1\% & 74.7\% & 45.8 \\
\ +FR$^\ast$ ($r=-2.0$) & $\tcbhighmath{\uparrow 4.5\%}$ 23.1\% & $\tcbhighmath{\uparrow 12.0\%}$ 37.3\% & $\tcbhighmath{\downarrow 44.4\%}$ 9.9\% & $\tcbhighmath[colback=LightOrange]{\downarrow 1.4\%}$ 27.7\% & $\tcbhighmath{\uparrow 0.3\%}$ 74.9\% & $\tcbhighmath[colback=LightOrange]{\downarrow 2.6\%}$ 44.6 \\
\ +FR$^\ast$ ($r=-4.0$) & $\tcbhighmath{\uparrow 8.6\%}$ 24.0\% & $\tcbhighmath{\uparrow 22.5\%}$ 40.8\% & $\tcbhighmath{\downarrow 57.9\%}$ {\bf 7.5\%} & $\tcbhighmath[colback=LightOrange]{\downarrow 4.3\%}$ 26.9\% & $\tcbhighmath[colback=LightOrange]{\downarrow 0.8\%}$ 74.1\% & $\tcbhighmath[colback=LightOrange]{\downarrow 6.1\%}$ 43.0 \\
\midrule 
\multicolumn{7}{c}{Rejection Learning \& Regularized Decoding (Ours)} \\
\midrule
\makecell{\textsc{RejL} ($\lambda=0.0$)} & 26.1\% & 39.1\% & 13.5\% & 28.1\% & 75.6\% & 45.1 \\
\ +RD ($\lambda=1.0$) & $\tcbhighmath{\uparrow 7.3\%}$ 28.0\% & $\tcbhighmath{\uparrow 5.9\%}$ 41.4\% & $\tcbhighmath{\downarrow 10.4\%}$ 12.1\% & $\tcbhighmath[colback=LightOrange]{\downarrow 1.1\%}$ 27.8\% & $\tcbhighmath{\uparrow 0.3\%}$ 75.8\% & $\tcbhighmath[colback=LightOrange]{\downarrow 1.6\%}$ 44.4 \\
\ +RD ($\lambda=2.0$) & $\tcbhighmath{\uparrow 12.3\%}$ 29.3\% & $\tcbhighmath{\uparrow 9.0\%}$ {\bf 42.6\%} & $\tcbhighmath{\downarrow 14.8\%}$ 11.5\% & $\tcbhighmath{\uparrow 0.0\%}$ 28.1\% & $\tcbhighmath{\uparrow 0.8\%}$ 76.2\% & $\tcbhighmath[colback=LightOrange]{\downarrow 3.8\%}$ 43.4 \\
\ +RD ($\lambda=3.0$) & $\tcbhighmath{\uparrow 16.5\%}$ {\bf 30.4\%} & $\tcbhighmath{\uparrow 8.2\%}$ 42.3\% & $\tcbhighmath{\downarrow 11.9\%}$ 11.9\% & $\tcbhighmath{\uparrow 1.1\%}$ {\bf 28.4\%} & $\tcbhighmath{\uparrow 1.1\%}$ {\bf 76.4\%} & $\tcbhighmath[colback=LightOrange]{\downarrow 5.8\%}$ 42.5 \\
\ +RD ($\lambda=5.0$) & $\tcbhighmath{\uparrow 23.8\%}$ {\bf 32.3\%} & $\tcbhighmath{\uparrow 11.0\%}$ {\bf 43.4\%} & $\tcbhighmath{\downarrow 20.0\%}$ 10.8\% & $\tcbhighmath{\uparrow 3.2\%}$ {\bf 29.0\%} & $\tcbhighmath{\uparrow 1.7\%}$ {\bf 76.9\%} & $\tcbhighmath[colback=LightOrange]{\downarrow 8.2\%}$ 41.4 \\
\bottomrule
\end{tabular}
\end{adjustbox}
\caption{\label{tab:auto_eval} Comparison between our proposed method and five baseline models. 
Numbers that are highlighted \tcbox{blue} represents the percentage of improvement and \tcbox[colback=LightOrange]{orange} denotes the percentage of performance drop. All models are trained on the \textsc{XSum} training set and
evaluated on 3K samples from the \textsc{XSum} test set. For MLE, we use the BART \cite{lewis-etal-2020-bart} model. SLT and TLT stands for sentence- and token-level loss truncation respectively. $c$ is the percentage of data to be dropped. In this experiment, all summaries are generated using beam search with beam size equals to 6. $\ast$: results reported in the original papers. $\ddagger$: we run evaluation ourselves using checkpoints the authors provided. $^\dagger$: model checkpoints are not available and we train the models ourselves using the author's code. Training hyper-parameters are based on the original paper.
}
\end{table*}
The findings of the factuality evaluation on the \textsc{XSum} test set are presented in Table \ref{tab:auto_eval}. We evaluate the effect of rejection loss without regularized decoding by setting $\lambda=0$. We make the following observations:

(1) {\bf In the evaluation of sentence-level summary factuality, our proposed method -- the rejection loss objective combined with regularized decoding -- outperforms all baseline methods.} Compared with MLE training, we observe a 41.7\% improvement (22.8\% $\rightarrow$ 32.3\%) on \textsc{FactCC} score and 25.4\% improvement (34.6\% $\rightarrow$ 43.4\%) on DAE score (when $\lambda=5.0$). Our method also outperforms MLE and the loss-truncation baseline in terms of entity-level factuality evaluation, but it underperforms the RL+FR baseline. We speculate that this is because the RL method is specifically optimized toward the entity-level factuality rewards generated using \textsc{EntFA}.

(2) {\bf We observe (almost) consistent improvement of factuality scores as we increase the value of $\lambda$.} This is in line with what we expected, as candidate summaries with uncertainty will be penalized more severely with a larger $\lambda$. An intriguing finding is that summary factuality is substantially increased when the rejection loss objective is used alone without regularized decoding. Compared to the MLE baseline, the summarization model trained with the rejection loss objective (without regularized decoding) achieves the same ROUGE-1 score 45.1 but significantly higher factuality scores ({\small \textsc{FactCC}}: $22.8\% \rightarrow 26.1\%$; {\small DAE}: $34.6\% \rightarrow 39.1\%$; {\small \textsc{EntFA}}: $15.5\% \rightarrow 13.5\%$) and has more novel $n$-grams. These findings imply that when the underlying dataset is noisy, the rejection loss objective performs generally better than the native cross-entropy objective for text summarization. It is unclear whether this generalizes to other text generation tasks, such as knowledge-grounded dialogue generation,  where training datasets frequently contain noise \cite{dziri2022origin}. This will be left for future work.

(3) {\bf The resulting summaries become more abstractive when $\lambda$ increases.} This is supported by the observation from Table \ref{tab:auto_eval}, where the generated summaries contain more novel $n$-grams as $\lambda$ increases. Thus, our method can simultaneously increase the abstractiveness and factuality of the generated summaries.  In contrast, the proportion of novel $n$-grams falls if we try to make the generated summaries more factual in loss truncation and RL methods.
The drop of novel $n$-grams suggests that factuality improvement brought by the baselines may come from increased level of extractiveness.
This is consistent with the findings of \citet{ladhak-etal-2022-faithful}. 

\begin{figure}
\centering
\begin{subfigure}[b]{\columnwidth}
   \includegraphics[width=0.94\textwidth]{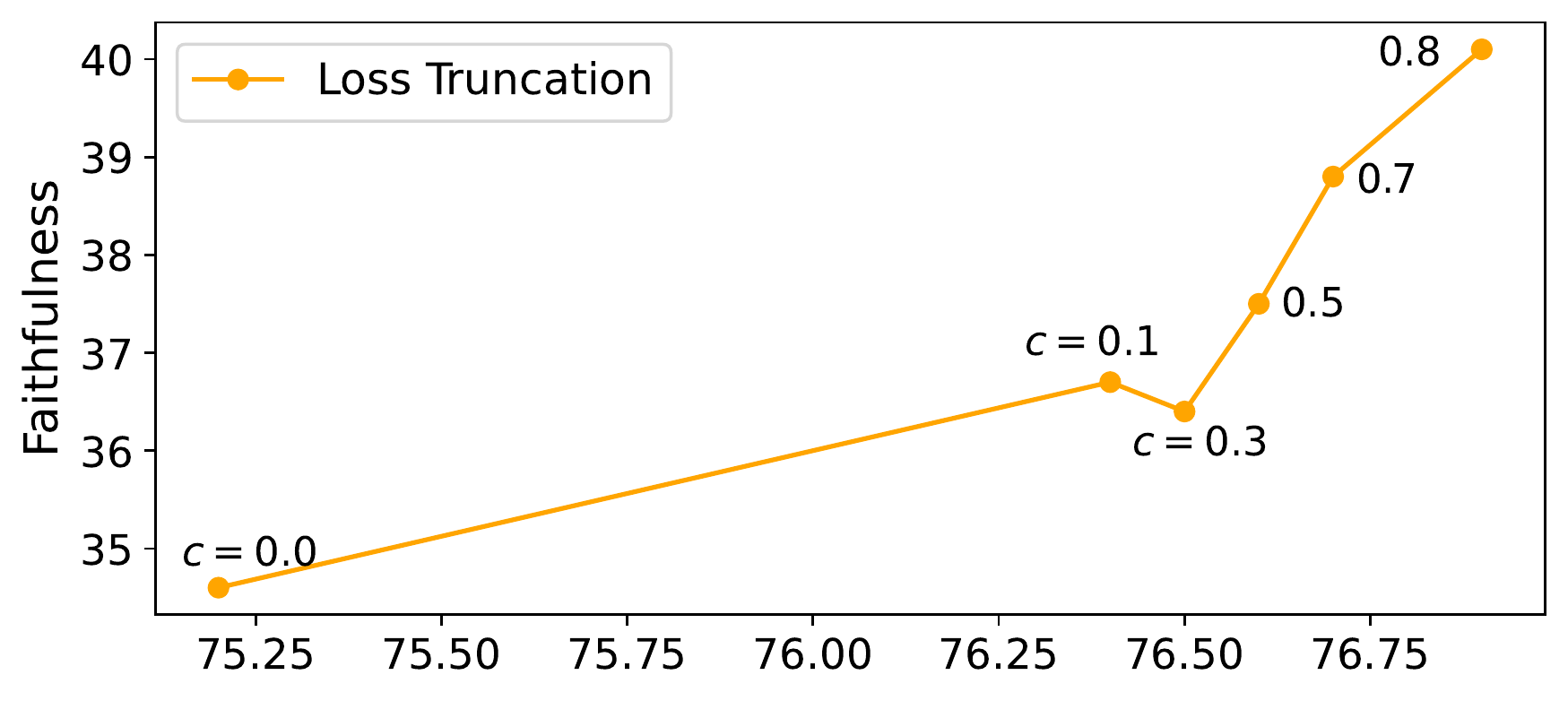}
   \label{fig:trade_off_1} 
\end{subfigure}
\begin{subfigure}[b]{\columnwidth}
   \includegraphics[width=0.94\textwidth]{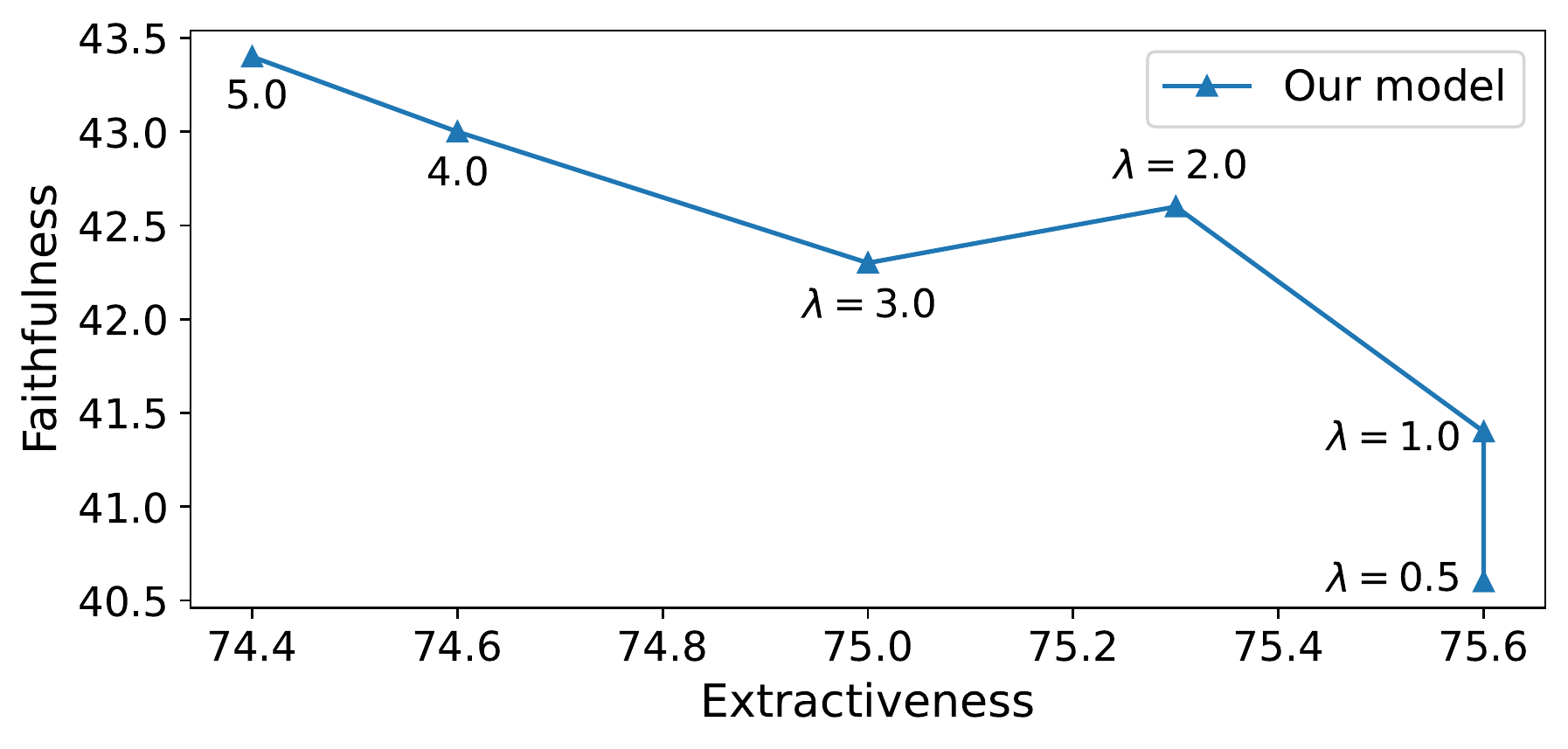}
   \label{fig:trade_off_2}
\end{subfigure}
\caption{Faithfulness-Abstractiveness trade-off curves for our model and the loss truncation baseline \cite{kang-hashimoto-2020-improved}. Extractiveness is measured using \emph{coverage} defined in \cite{grusky-etal-2018-newsroom}. Faithfulness is measured using the DAE score \cite{goyal-durrett-2021-annotating}.}
\label{fig:trade_off}
\end{figure}

We also plot the Faithfulness-Abstractiveness Tradeoff curve following the recent work by \citet{ladhak-etal-2022-faithful}, as seen in Figure~\ref{fig:trade_off}. We adopt \emph{coverage} defined in \citet{grusky-etal-2018-newsroom} to measure the level of extractiveness: a summary with higher coverage is considered more extractive. Faithfulness is measured using DAE score. In Figure~\ref{fig:trade_off}, the abstractiveness and factuality of our model are positively correlated. The loss truncation approach, in contrast, exhibits a negative correlation. This result further confirms our point.

(4) {\bf We observe a decline in ROUGE scores as the summary factuality increases.} According to our calculation, the average length of the generated summaries for different $\lambda$ remains constant (around eighteen words). Hence,  it is unlikely that summary length played a role in the declining ROUGE score.  Instead, we speculate that the decline in the ROUGE scores is caused by the increase in abstractiveness. For instance, the regularized decoded summary in Table \ref{tab:example} is highly abstract.  It makes no mention about the funding, as the value of the fund is not specified in the source. 
In contrast, the summary generated by BART matches the reference more closely in terms of $n$-gram overlap even though it contains non-factual hallucinations. In light of this example, it appears that the MLE objective may encourage the summarization model to hallucinate in order to boost ROUGE scores.

\subsubsection{Human Evaluation}
\begin{table}[t!]
\renewcommand{\arraystretch}{0.93}
\begin{center}
\begin{adjustbox}{width=\columnwidth, center}
\begin{tabular}{l|c|c}
\toprule
~ &  vs. MLE & vs. Loss Trunc. \\
\midrule
Win & 39.3\% & 32.0\% \\
Lose & 23.3\% & 26.0\% \\
Equally Factual & 17.3\% & 22.7\% \\
Equally Non-factual & 20.0\% & 19.3\% \\
\bottomrule
\end{tabular}
\end{adjustbox}
\end{center}
\caption{\label{table:human_eval} Human comparisons on 150 randomly selected articles from \textsc{XSum} test set. Win: percentage of outputs from our model that are preferred by the annotators over the baseline's outputs, given the same source document.}
\end{table}
In addition to automatic evaluation, we also conducted a human evaluation, comparing our model ($\lambda$=2.0) to the MLE and the token-level loss truncation $c=0.5$ baseline. We showed participants the output of our model and a comparison model, and asked them to make a pairwise preference judgment between the two summaries based on their faithfulness. 

Specifically, we randomly sampled 150 articles from the \textsc{XSum} test set for summarization. To make the evaluation process more effective, we filtered out samples where the generated summary pair has more than 70\% unigram overlap. We presented the two summaries in random order and ask participants to carefully consider the faithfulness of each summary. We then asked them to select the summary that they prefer, or to say they are equally factual/non-factual. A more detailed annotation guide can be found in Appendix \ref{sec:guideline}.

Table~\ref{table:human_eval} shows the human evaluation results. From the table, we can find that summaries generated by our model are preferred by the human annotators over two baselines.\footnote{Discussion of inter-annotator agreement is included in Appendix \ref{sec:inter_agreeemtn}.}

\section{Analysis}
\subsection{The Rejection Probability of Hallucinated Tokens}\label{sec:rej_analysis}
\begin{table}[t!]
\renewcommand{\arraystretch}{1.1}
\begin{center}
\begin{adjustbox}{width=\columnwidth, center}
\begin{tabular}{rlr}
\toprule
 \multicolumn{3}{l}{The University of Manchester has been awarded \_\_\_} \Bstrut \\ 
\Tstrut{\bf MLE:} & \multicolumn{2}{l}{\tcbox[colback=LightOrange]{\pounds 3.5m} to help fund a new research centre.} \Bstrut \\ 
\hline
\Tstrut{\bf Ours:} & \pounds 3.5m (...) & \% same ent. \\ 
~ & \pounds 1.5m (...) & \% different hal. \\ 
~ & \pounds 12.8m (...) & \% factual \\ 
~ & a large amount of money to (...) \; & \% remove ent. \\ 
~ & \texttt{<REJ>} \; & \% rejection token \\ 
\bottomrule
\end{tabular}
\end{adjustbox}
\end{center}
\caption{\label{tab:ref_acc} An example of how we evaluate our rejection model on \textsc{XEnt} \cite{cao-etal-2022-hallucinated}. In this example, \tcbox[colback=LightOrange]{\pounds 3.5m} is generated by BART and labeled as non-factual hallucination by human annotator. We provide the same context to our model and analyze the distribution of five different types of outputs from out model.
We did this analysis for both factual and non-factual entities in \textsc{XEnt}.}
\end{table}

In this section, we analyze how well the rejection class calibrates with hallucinated entities.
We use the test set of \textsc{XEnt} dataset, which contains 240 machine generated summaries with entity-level hallucination annotations. Each entity is annotated by human experts for its hallucination and factuality status. 

For each annotated entity in \textsc{XEnt}, we give our model the same context as input and analyze the model's output. For non-factual entities, we expect the model will generate the rejection token. Table \ref{tab:ref_acc} shows the evaluation setup. In this example, ``\emph{The University of Manchester has been awarded £3.5m to help fund a new cancer research centre.}'' is one annotated sample from the \textsc{XEnt} test set. The entity ``\pounds3.5m'' is labeled as non-factual hallucination by human annotators.
We use the left context (i.e., the first row in Table \ref{tab:ref_acc}) as well as the source document as input to our model.
There are five possible outputs from our model: (1) the same entity (2) a different entity that is hallucinated and non-factual (3) a different entity this is factual (4) removing the entity and replace it with an abstract description (e.g., ``an 90-year-old woman'' $\rightarrow$ ``an elderly woman'') (5) a rejection token. The results are summarized in Table \ref{table:rej_align}.

\begin{table}[t!]
\renewcommand{\arraystretch}{0.95}
\begin{center}
\begin{adjustbox}{width=\columnwidth, center}
\begin{tabular}{l|c|c|c}
\toprule
~ & \multicolumn{2}{c|}{\bf Factual Entity} & \multirow{2}{*}{\bf Non-factual} \\
~ &  Non-hal. &  Factual Hal. & ~ \\ 
\midrule
Same Ent. & 81.8\% & 64.4\% & 17.2\% \\
Different Hal. & 0.0\% & 2.5\% & 0.7\% \\
Factual Ent. & 0.2\% & 0.0\% & 0.0\% \\
Remove Ent. & 5.1\% & 7.6\% & 4.5\% \\
Rejection & 13.0\% & 25.4\% & 77.6\% \\
\midrule
Total & 583 & 118 & 134 \\
\bottomrule
\end{tabular}
\end{adjustbox}
\end{center}
\caption{\label{table:rej_align} Distribution of five different types of outputs for different entities. {Non-hal.} stands for non-hallucinated factual entity. {Factual Hal.} are hallucination that cannot be directly entailed in their generated context from the source document but can be based on world knowledge.}
\end{table}

As shown in Table \ref{table:rej_align}, when the entity is factual, our model will not change the generation in most cases (78.9\%). When the entity is non-factual, however, 77.6\% of the time our model will generate the rejection token and 4.5\% of the time the hallucination will be removed. These results indicate that the rejection model works as we expect.
We also notice that the model will reject a factual entity about 15\% of the time. This number can be reduced by increasing the regularization weight $\alpha$ in Equation \ref{eq:obj}. For example, by increasing $\alpha$ from 1.0 to 2.0, the rejection rate for factual entities drops from 15\% to 6.8\%. However, the rejection rate for non-factual entities also drops to 68.7\%.

\subsection{The Effect of Beam Size on Factuality}
\begin{figure}[t!]
\centering
\includegraphics[scale=0.39]{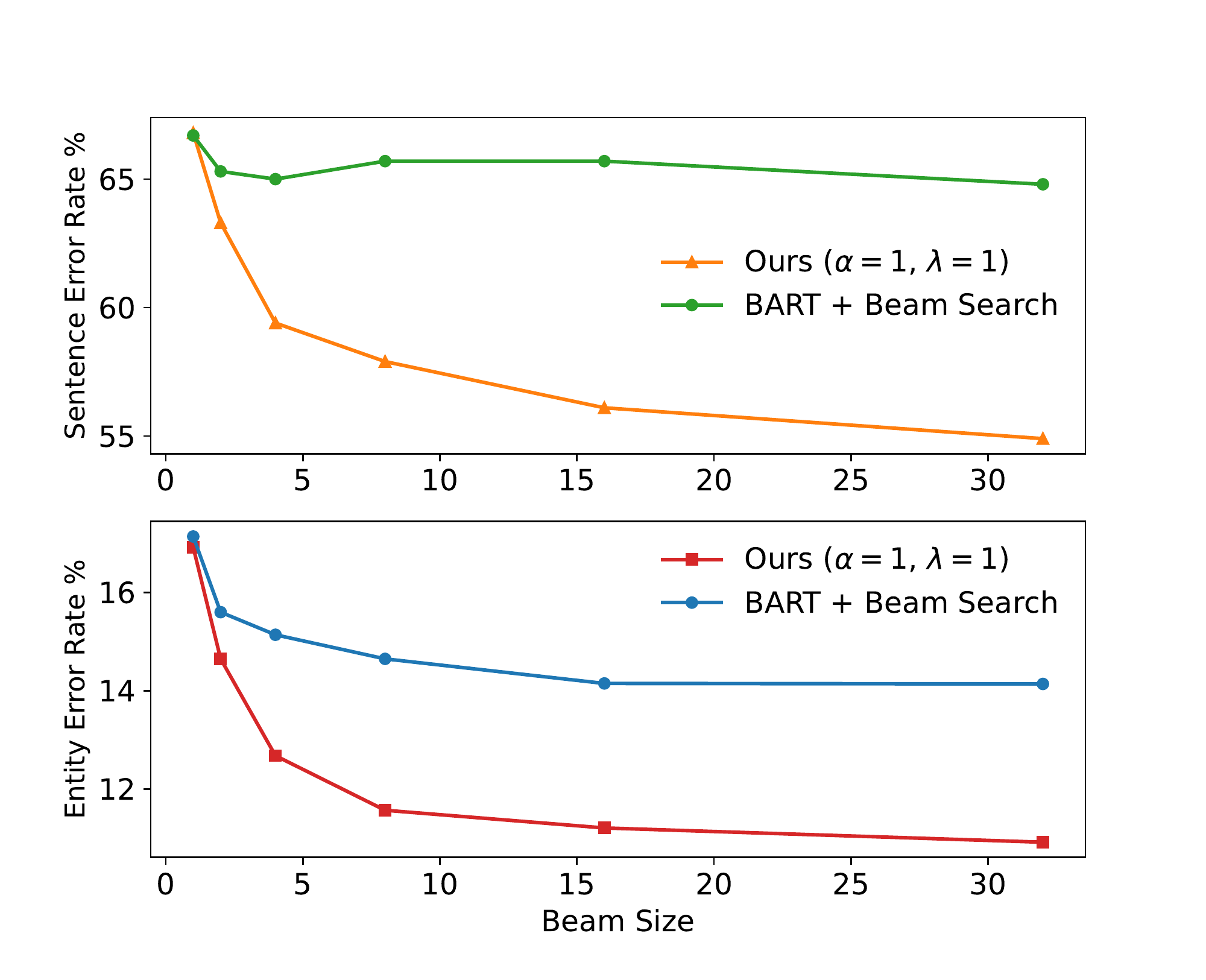}
\caption{Beam size and factuality. We calculate the sentence error rate using DAE \cite{goyal-durrett-2021-annotating} and non-factual entity hallucination rate using \textsc{EntFA} \cite{cao-etal-2022-hallucinated}.}
\label{figure:analysis_bs}
\end{figure}
Beam search is widely used to decode sentences from neural text generation models. Previous work has shown that increasing the beam size (i.e., the number of beam candidates held during searching) does not necessarily improve the model's performance \cite{stahlberg-byrne-2019-nmt}. However, little work has been done to investigate the effects of beam size on summary factuality. In this section, we evaluate the factuality of summaries generated using different beam size. 

We compare the factuality of our model (rejection loss + regularized decoding) and the pre-trained BART model when decoded with different beam sizes. We select the beam sizes by powers of 2 from 1 to 32.
In Figure \ref{figure:analysis_bs}, we observe that the factuality of summaries generated by BART slightly improves as the beam size increases. Our model, on the other hand, is able to achieve much better factuality score with a larger beam size. 
It is worth noting that our method has almost the same error rate as the baseline when doing greedy decoding (i.e., $\textrm{beam size}=1$). Then the error rate decreases rapidly as the beam size increases. 
These results demonstrate that our method has a better inductive bias to identify non-factual beam candidates and remove them.


\section{Conclusion}
In this work, we propose a rejection learning based training objective for abstractive summarization models that is robust to the presence of noise in the dataset. The proposed training objective allows the model to learn features that are associated with the noise and reject them during training.
We additionally propose a novel regularized decoding objective that effectively penalizes non-factual candidate summaries during inference.
We show that our method vastly improves the factuality of generated summaries over five baseline models using both automatic and human evaluation.
Further analysis reveals that our model enhances the factuality of the summaries while simultaneously increasing the abstractiveness of the generated summaries. 


\section{Limitations}
Our assumption in this work is that noisy references contain information that is absent from the source, which creates an impossible learning task for the summarization model. As a result, noisy samples have high loss during training, which allows rejection learning to identify and reject them. However, it is not clear whether there also exist ``difficult'' but clean samples that also have high loss because they are simply difficult for the model to learn.
For example, some articles in the training set contain numerical data, and arithmetic reasoning may be required to generate a good summary. If such samples exist, it is possible that our method would reject them instead of learning. Whether this happens remains to be evaluated.

In addition, the proposed method is only tested on the summarization task. Although our method can be applied to other NLG tasks such as machine translation and knowledge-grounded dialogue generation, it remains to be seen whether a similar advantage in factuality can be obtained.


\bibliography{anthology,custom}
\bibliographystyle{acl_natbib}
\appendix
\clearpage
\section{Annotation Guideline}\label{sec:guideline}
We conduct human evaluation based on randomized pair comparison. With the following guidelines, we present annotators with 1) the source 2) and two systems' outputs in random order:
\begin{enumerate}
    \item Read the source completely.
    \item Based on the source, compare the generations of system 1 with system 2 in terms of level of ``factuality'' with respect to the source document.
    \item Try to consider the following questions when doing factuality evaluation:
    \begin{itemize}
        \item Does the summary accurately capture the important information in the original document?
        \item Is the salient information in each summary, such as location, date, quantity, etc., consistent with the original document?
        \item Does the generated summary contain information that cannot be directly inferred from the source document? (avoid using external knowledge unless it is common sense)
        \item Does the summary unfoundedly expand the meaning of the source document? 
        \item Are the relationship between characters, the sequence of events, and the causal relationship in the summary consistent with the original text?
    \end{itemize}
    \item Select the summary that you think is more faithful with respect to the source document (i.e., contain less unfounded facts, more accurate). Or they are equally factual/non-factual.
\end{enumerate}

\section{Inter-Annotator Agreement} \label{sec:inter_agreeemtn}
We create a common set of 50 samples for inter-annotator agreement evaluation. We ask three annotators to annotate the common set. We report Fleiss's Kappa ($\kappa$) to assess the reliability of agreement between annotators. On the four-category annotation, we obtain a fair agreement with $\kappa = 0.309$ ($0.20 \leq \kappa \leq 0.40$). We also report the percentage $\mu$ of annotators that annotate the majority class for the given example \cite{durmus-etal-2020-feqa}. We get $\mu = 0.633$ on our four-category annotation.

We speculate that the reason for the relatively low Fleiss's Kappa score is because of the introduction of the \emph{equally factual/non-factual} class which makes the annotation task hard. For example, when the two summaries both contain errors, some annotators prefer to select the one with less errors (as we specified in the guideline) but some annotators tend to mark both summaries as equally non-factual. When all three annotators only annotate win or loss, we get a substantial agreement with $\kappa = 0.615$ ($0.61 \leq \kappa \leq 0.80$) and $\mu = 0.912$.

\section{Max Regularizer} \label{sec:appendix_maxreg}
\begin{table}[t!]
\renewcommand{\arraystretch}{0.90}
\begin{center}
\begin{adjustbox}{width=\columnwidth, center}
\begin{tabular}{l|c|c|c|c}
\toprule
~ & \multicolumn{2}{c|}{\bf Factuality} & \multirow{2}{*}{\bf ROUGE-1} & \multirow{2}{*}{\bf \makecell{Novel \\ Unigram}} \\
~ & {\small \textsc{FactCC} $\uparrow$} &  {\small \textsc{DAE} $\uparrow$} & ~ & ~ \\ 
\midrule
$\lambda=1.0$ & 26.8\% & 40.0\% & 44.7 & 27.9\% \\
$\lambda=3.0$ & 28.3\% & 41.0\% & 44.4 & 27.9\% \\
$\lambda=5.0$ & 28.5\% & 41.2\% & 43.9 & 28.1\% \\
$\lambda=10.0$ & 30.0\% & 41.7\% & 42.7 & 28.6\% \\
\bottomrule
\end{tabular}
\end{adjustbox}
\end{center}
\caption{\label{tab:max_reg}Evaluation results on \textsc{XSum} test set using the max regularizer proposed in Section \ref{sec:regularzied_decoding} (Equation \ref{eq:max_reg}). We set $\alpha=1.0$.}
\end{table}

Table \ref{tab:max_reg} shows the evaluation results when we use the max regularizer defined in Section \ref{sec:regularzied_decoding}. Compared with the sum regularizer (Equation \ref{eq:sum_reg}), we find that the max regularizer performs slightly worse in terms of automatic factuality evaluation scores. Thus we use the sum regularizer in our main results.

\section{Example Appendix} \label{sec:appendix_examples}
\begin{table*}[t]
\renewcommand{\arraystretch}{1.15}
\centering
\small
\begin{tabular}{p{15.5cm}}
  \toprule
  \Tstrut{\bf Source}: \\
    St Michael's Hospital is joining other UK foetal medicine centres to set up a twin-to-twin transfusion syndrome (TTTS) registry to share information. Consultant Mark Denbow said they were ``constantly learning'' about TTTS. He said the ``rare and often devastating condition'' occurs in about 10-15\% of identical twin pregnancies. St Michael's Hospital is also one of a few UK centres offering laser ablation surgery, where doctors can operate on the babies while they are in the womb. Jo and Finbarr O'Halloran, from Backwell, said they were ``devastated'' when they discovered  their twins had the syndrome during a hospital scan. Mrs O'Halloran was given laser ablation surgery at about 22 weeks into the pregnancy in 2013. A few weeks later, daughters Eve and Amy were born by emergency caesarean section. Jo O'Halloran said: ``I am literally thankful every single day that they survived and that they're here.'' Keith Reed, from the Twins and Multiple Birth Association (Tamba), said they were ``still only skimming the surface'' in terms of data collection. He said: ``In order to build the best possible picture of TTTS cases in the UK, and help see which treatments offer the best possible outcomes, we need more hospitals with foetal medicine departments to sign up.'' TTTS affects twins who share a placenta, where the babies are not sharing blood equally. One baby gets too much blood and the other baby does not get enough. If left untreated 90\% of these babies will die and even with treatment there is only up to 70\% chance of both babies surviving. \Bstrut \\
  \hline
  \Tstrut{\bf Reference}: \\
  A study to improve the survival rate of unborn twins, with a potentially life threatening syndrome, is under way in Bristol. \Bstrut \\
  \hline
  \Tstrut{\bf Baseline BART}: \\
  A \tcbox[colback=LightOrange]{Swansea} hospital is to set up a registry of twin-to-twin transfusion syndrome. \\
  \hline
  \Tstrut{\bf Baseline Sentence Loss Truncation ($c=0.5$)}: \\
  A \tcbox[colback=LightOrange]{Worcestershire} hospital has become \tcbox[colback=LightOrange]{the first} in the UK to set up a registry for twin transfusion syndrome. \\
  \hline
  \Tstrut{\bf Ours}: \\
  A new registry is being set up to help save the lives of babies born with a rare blood disorder. \Bstrut \\
  \hline
  ~ \\
  \hline
  \Tstrut{\bf Source}: \\
  The 30-year-old is to remain there following the completion of a psychiatric report. Belfast Magistrates' Court was told she would not be able to leave without the hospital's permission. The woman cannot be named amid claims that identifying her would increase the risk of her taking her own life. A press challenge to the temporary reporting restrictions is due to be heard next month. The woman was arrested by detectives investigating the child's death following an incident in Belfast in March. She was charged with murder and then held under the Mental Health Act. Her barrister revealed on Wednesday that a medical report had now been prepared. Based on its contents he sought a termination of the current arrangements for keeping his client at the facility. ``The application is that she be remanded on bail, subject to the condition that she continues to reside (there),'' he said. A doctor who assessed the accused confirmed that the health trust consented to the proposal, provided the accused was there as a detained person. The judge was informed that under those arrangements the woman would not be able to leave. Granting the application, she listed the criminal proceedings for a further update in eight weeks time.
  \Bstrut \\
  \hline
  \Tstrut{\bf Reference}: \\
  A woman accused of murdering her baby son has been remanded on bail at a mental health facility. \Bstrut \\
  \hline
  \Tstrut{\bf Baseline BART}: \\
  A woman charged with murdering her \tcbox[colback=LightOrange]{two-year-old daughter} has been remanded in custody at the \tcbox[colback=LightOrange]{Royal Victoria Hospital} in Belfast. \\
  \hline
  \Tstrut{\bf Baseline Sentence Loss Truncation ($c=0.5$)}: \\
  A woman charged with murdering her \tcbox[colback=LightOrange]{two-year-old daughter} has been remanded in custody at Belfast's \tcbox[colback=LightOrange]{Royal Victoria Hospital}. \\
  \hline
  \Tstrut{\bf Ours}: \\
  A woman charged with murder has been remanded in custody after a judge granted her bail on the condition that she remains in hospital. \Bstrut \\
  \bottomrule
\end{tabular}
\caption{Comparison of summaries generated by the standard BART model, BART trained with sentence loss truncation and BART trained using our method. The factual errors made by the models are highlighted \tcbox[colback=LightOrange]{orange}.
}
\end{table*}

\end{document}